\documentclass[conference]{IEEEtran}
\IEEEoverridecommandlockouts
\usepackage{cite}
\usepackage{amsmath,amssymb,amsfonts}
\usepackage{algorithmic}
\usepackage{graphicx}
\usepackage{textcomp}
\usepackage{xcolor}
\usepackage{hyperref}
\usepackage{booktabs,multirow}
\usepackage[linesnumbered,ruled,vlined]{algorithm2e}
\usepackage{caption}
\def\BibTeX{{\rm B\kern-.05em{\sc i\kern-.025em b}\kern-.08em
    T\kern-.1667em\lower.7ex\hbox{E}\kern-.125emX}}
\begin{document}

\title{Scribble-Supervised RGB-T Salient Object Detection\\
\thanks{*Corresponding author. This paper is supported by Natural Science Foundation of Anhui
Province (2108085MF210), Key Program of Natural
Science Project of Educational Commission of Anhui Province
(KJ2021A0042).}
}

\author{\IEEEauthorblockN{1\textsuperscript{st} Zhengyi Liu}
\IEEEauthorblockA{\textit{Computer Science and Technology} \\
\textit{Anhui University}\\
Hefei, China\\
liuzywen@ahu.edu.cn}
\and
\IEEEauthorblockN{2\textsuperscript{nd} Xiaoshen Huang}
\IEEEauthorblockA{\textit{Computer Science and Technology} \\
\textit{Anhui University}\\
Hefei, China \\
1045258695@qq.com}
\and
\IEEEauthorblockN{3\textsuperscript{rd} Guanghui Zhang}
\IEEEauthorblockA{\textit{Computer Science and Technology} \\
\textit{Anhui University}\\
Hefei, China\\
2532950974@qq.com}
\and
\IEEEauthorblockN{4\textsuperscript{th} Xianyong Fang}
\IEEEauthorblockA{\textit{Computer Science and Technology} \\
\textit{Anhui University}\\
Hefei, China\\
fangxianyong@ahu.edu.cn}
\and
\IEEEauthorblockN{5\textsuperscript{th} Linbo Wang*}
\IEEEauthorblockA{\textit{Computer Science and Technology} \\
\textit{Anhui University}\\
Hefei, China\\
wanglb@ahu.edu.cn}
\and
\IEEEauthorblockN{6\textsuperscript{th} Bin Tang}
\IEEEauthorblockA{\textit{Artificial Intelligence and Big Data} \\
\textit{Hefei University}\\
Hefei, China\\
424539820@qq.com}
}

\maketitle

\begin{abstract}
Salient object detection segments attractive objects in scenes. RGB and thermal modalities provide complementary information and scribble annotations alleviate large amounts of human labor. Based on the above facts, we propose a scribble-supervised RGB-T salient object detection model. By a four-step solution (expansion, prediction, aggregation, and supervision), label-sparse challenge of scribble-supervised method is solved. To expand scribble annotations, we collect the superpixels that foreground scribbles pass through in RGB and thermal images, respectively. The expanded multi-modal labels provide the coarse object boundary. To further polish the expanded labels, we propose a prediction module to alleviate the sharpness of boundary. To play the complementary roles of two modalities, we combine the two into aggregated pseudo labels. Supervised by scribble annotations and pseudo labels, our model achieves the state-of-the-art performance on the relabeled RGBT-S dataset. Furthermore, the model is applied to RGB-D and video scribble-supervised applications, achieving consistently excellent performance.\href{https://github.com/liuzywen/RGBTScribble-ICME2023}{https://github.com/liuzywen/RGBTScribble-ICME2023.}
\end{abstract}


\section{Introduction}
\textbf{Background.}
%Salient object detection (SOD) aims to segment the objects that attract human attention most in a scene.
In the cluttered background, low-illumination, and foggy
weather scenes, the thermal image is a promising supplement to the RGB image because it is insensitive to lighting and weather by capturing the radiated heat of objects. Therefore, RGB-T salient object detection (SOD) which segments  attractive objects shows promising research prospect.%has drawn increasing attention.}}

%The thermal image can capture the radiated heat of objects, and it is insensitive to lighting and weather conditions, and suitable for handling scenes captured under adverse conditions, for example, total darkness environment, foggy weather, and cluttered backgrounds.

%Scribble annotation supervised methods which can %both reduce the labeling cost and guarantee the detection accuracy
%achieve a trade-off between labeling efficiency and model performance
%have

%DENet\cite{xu2022weakly} and WSVOD \cite{zhao2021weakly} introduce the first scribble  supervised RGB-D and video SOD datasets, respectively. We construct the scribble  supervised RGB-T ones. Based on the three multi-modal scribble supervised datasets (Fig \ref{fig:Dataset}(a-e)), we propose multi-modal SOD method via scribble supervision.

\textbf{Challenge.}
Scribble annotations can alleviate the labeling cost and have been widely used in remote sensing image \cite{wei2021scribble,li2022progressive}, medical image \cite{luo2022scribbleseg,chen2022scribble2d5}, and 3D image segmentation \cite{unal2022scribble}.  However, object structures and boundary details cannot be easily inferred \cite{zhang2020weakly}. %because scribble annotations provide sparse information only.
%It is the main challenge in scribble-supervised methods.
Especially, it is more difficult to extract the complete objects from low-quality RGB images.

\begin{figure}[!htp]
   \center
  \includegraphics[width=1\linewidth]{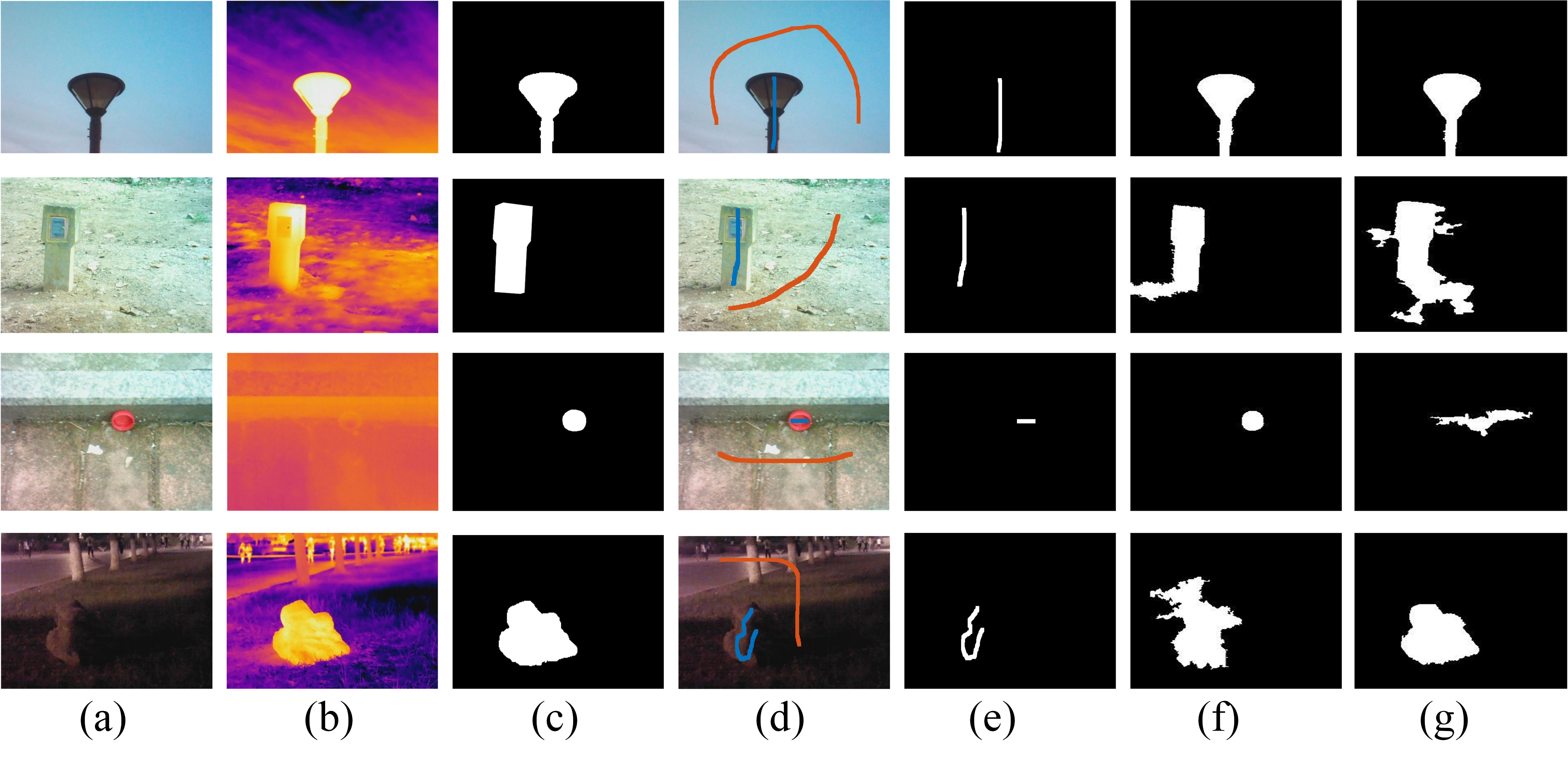}
  %\vspace{-2em}
  \caption{{Scribble-supervised RGB-T SOD datasets and corresponding multi-modal expanded labels. (a) RGB images (b) thermal images (c) ground truth (d) scribble annotations ({\color{red}{red}}: background; {\color{blue}{blue}}: foreground) (e) foreground scribbles (f) RGB expanded  labels (g) thermal expanded labels.}
  \label{fig:Dataset}}
\end{figure}

\textbf{Existing Solutions.}
Some methods \cite{xu2022weakly,zhao2021weakly,gao2022weakly} use edge detection to provide more structure information.
Some methods \cite{wei2021scribble,chen2022scribble2d5} use simple linear iterative clustering (SLIC) \cite{achanta2012slic} to collect the superpixels in RGB images that scribbles pass through, generating pseudo labels for supervision. Some methods \cite{xu2022weakly,gao2022weakly} solve the challenge of low-quality RGB images by appending depth and optical flow information.
%Some methods\cite{wei2021scribble,chen2022scribble2d5} use simple linear iterative clustering (SLIC)\cite{achanta2012slic} to segment RGB images according to image intensity and distance similarities and collect the superpixels that scribbles pass through, resulting in psuedo labels.
%The enlarged pseudo labels are effective in image segmentation tasks  due to the larger supervision regions compared with scribbles.

\textbf{Our Findings.} Scribble annotations  describe the partial cues of objects without complete boundary, while superpixel segmentation can provide the coarse object boundary according to the local consistency in features.
As a result, according to superpixels, we expand the foreground scribble annotations to generate the expanded  labels from the two modalities (RGB and Thermal).
By the observations from Fig \ref{fig:Dataset} (f-g), we find  that expanded  labels provide abundant and useful information($1^{st}$ row). Furthermore, we also find that expanded  labels are inaccurate due to some noisy boundaries ($2^{nd}$ row). Most importantly, the complementarity of two modalities is disclosed that the expanded  labels from one modality are inaccurate while those from the other modality match the salient objects ($3^{rd}$ and $4^{th}$ rows).
Accordingly, besides original scribble supervision, we append the supervision of the expanded  labels.
However, the expanded labels are not always accurate due to the limitations of the superpixel segmentation, which will affect the final prediction.

\textbf{Our Solution.}
The four-step solution,
\textit{Expansion}, \textit{Prediction}, \textit{Aggregation}, and \textit{Supervision},
%\textit{Expansion}$\rightarrow$\textit{Prediction}$\rightarrow$\textit{Aggregation}$\rightarrow$ \textit{Supervision},
is proposed to solve the above problem. Specifically, we first use RGB and thermal superpixels to expand foreground scribble annotations, generating the multi-modal expanded  labels. Then,
%two prediction modules are used to predict the multi-modal pseudo labels. Next, The multi-modal predicted pseudo labels are aggregated into a final pseudo label which is used to supervise model training  based on original scribble supervision.
we propose a prediction module to predict the expanded  labels from two modalities, respectively. Furthermore, the predicted multi-modal labels are averaged and smoothed to form the aggregated pseudo labels which can be further used to supervise the model training.

\textbf{Our Contributions.}
%\vspace{0.04em}
\begin{itemize}
%\vspace{-0.9em}
    \item
    %For the \textbf{first} time, we launch a scribble-supervised RGB-T SOD {\color{blue}dataset}, named RGBT-S. Based on the  RGBT-S, a scribble-supervised RGB-T SOD model is proposed. The performance is near to that of the state-of-the-art {\color{blue}{fully-supervised}} methods.
    A scribble-supervised RGB-T SOD model is proposed based on our launched scribble-supervised RGBT-S dataset, achieving the comparable performance with the state-of-the-art fully-supervised methods. Multi-modal superpixel expansion  obtains the completeness of objects by the combination of RGB and thermal images, so that the excellent pseudo labels are generated.% The performance is near to that of the state-of-the-art fully-supervised methods.
%\vspace{-0.9em}
    \item
    Different from existing superpixel expansion models  (ScRoadExtractor \cite{wei2021scribble}, Scribble2D5 \cite{chen2022scribble2d5}), we  propose to supervise the model via the predicted pseudo labels instead of direct expanded ones. %The predicted labels can polish the inaccurate boundaries to reduce the errors.
%\vspace{-0.9em}
    \item
    Different from existing multi-modal scribble-supervised SOD models  (RGB-D DENet \cite{xu2022weakly}, Video WSVOD \cite{zhao2021weakly}), we achieve multi-modal complementarity exploration in the supervision, rather than only in the model structures.

\end{itemize}

\section{Proposed method}
%\subsection{Motivation}
%Scribble annotations only describe the partial cues of objects without complete boundary, while superpixel segmentation can provide the coarse object boundary according to the local consistency in features. Therefore, we expand the foreground scribble annotations according to superpixels to generate the expanded labels. From Fig \ref{fig:Dataset}, we can find some expanded labels match with the ground truth, some expanded labels have noisy  boundaries, and the expanded labels from two modalities have the complementary role. As a result, besides original scribble supervision, we append the supervision of the expanded labels. However, the expanded labels are not always accurate due to the limitations of the superpixel segmentation. Therefore, we propose a prediction module to predict the expanded labels from two modalities, respectively. Furthermore, The predicted multi-modal expanded labels are averaged and smoothed to form the aggregated pseudo labels which can be further used to supervise the whole network.

\subsection{Overview}
\begin{figure*}[!htp]
\center
  \includegraphics[width=0.95\linewidth]{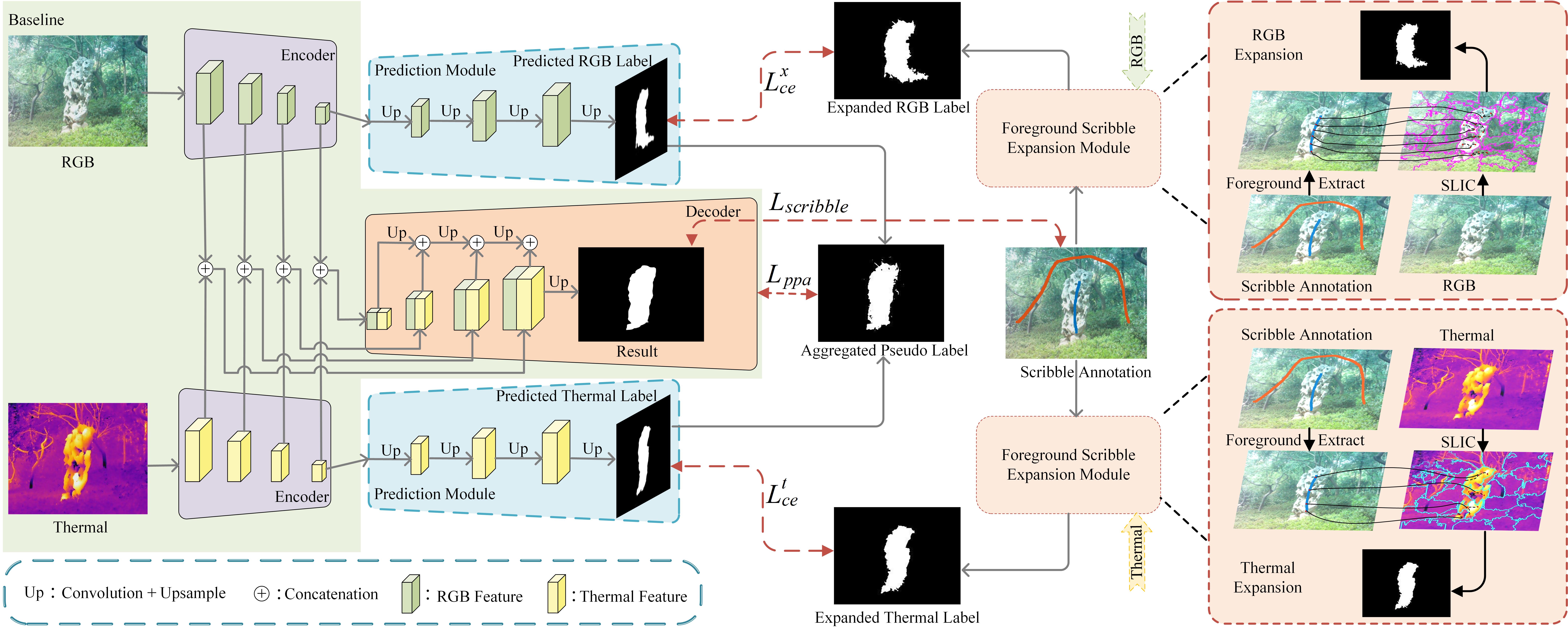}
  \caption{%\footnotesize
  {The framework of the proposed model. %The baseline network is a conventional two-stream network supervised by the scribble annotation. Next, the scribble annotation is expanded via RGB superpixels and thermal superpixels to generate the expanded labels. Due to the limitation of superpixel segmentation, the expanded labels are inaccurate. A prediction module is proposed to predict the expanded label. As a result, two predicted labels are obtained and then aggregated into a pseudo label which is used to supervise the model.
  }
  \label{fig:Main}}
\end{figure*}
%\vspace{-1.5em}
The model (Fig \ref{fig:Main}) builds on a baseline network supervised by scribble annotations. Then, two foreground scribble expansion modules are used to expand the annotations via the superpixels, generating the expanded multi-modal labels. Then, the expanded labels supervise the prediction modules to generate the predicted labels. Next, the predicted labels are aggregated into the pseudo labels which will additionally supervise the baseline model.

%Two foreground scribble \textbf{expansion} modules are used to generate the expanded multi-modal labels.
%Two \textbf{prediction} modules are used to generate the predicted multi-modal labels.
%An \textbf{aggregation} process is used to generate the aggregated pseudo labels.
%An additional \textbf{supervision} via the aggregated pseudo labels is used to train the model.

%Our training dataset is defined as $D=\{\{X\}_{i=1}^M,\{T\}_{i=1}^M,\{Y\}_{i=1}^M\}$, where $X=\{x_i\}^N_{i=1}$ is the RGB image set, $T=\{t_i\}^N_{i=1}$ is thermal image set, and $Y=\{y_i\}^N_{i=1}$ is scribble annotation, where $N$ is the number of pixels.
\subsection{Scribble-Supervised Baseline Model}

In the baseline network shown in the green background of Fig \ref{fig:Main}, RGB image $X$ and thermal image $T$ are fed into two encoders  %{\color{blue}{(ResNet \cite{he2016deep} or PVT \cite{wang2021pyramid})}}
to extract multi-modal features $F^x=\{f^x_i\}_{i=1}^{4}$ and $F^t=\{f^t_i\}_{i=1}^{4}$, respectively, where $i$ denotes the number of layers.
%Atrous Spatial Pyramid Pooling (ASPP) \cite{chen2017rethinking} modules which can enlarge the receptive fields are also used in the highest layer of each encoder.
Then these features are concatenated and fed into a  conventional saliency decoder which consists of adjacent layer concatenation, convolution, and upsampling operations to generate the result saliency map $R=\{r_i\}_{i=1}^N$ which is further supervised by the scribble annotation $Y=\{y_i\}_{i=1}^N$, where $y_i=1/2/0$ indicates foreground, background, and unknown pixels, respectively, and $N$ is the number of pixels in each image.

%Here, the encoder adopts ResNet and Pyramid Vision Transformer (PVT) \cite{wang2021pyramid}.

\begin{equation}
\begin{aligned}
F^x=\textit{Encoder}(X)\\
F^t=\textit{Encoder}(T)\\
R=\textit{Decoder}(\textit{Concat}(F^x,F^t))\\
\end{aligned}
\end{equation}
where $\textit{Concat}(\cdot)$ denotes concatenation operation.

%Then we give the loss functions of scribbles supervision.
Inspired by WSSA \cite{zhang2020weakly} and SCWSSOD \cite{yu2021structure}, we use (1) partial cross entropy loss ($L_{pce}$) to supervise predicted saliency pixels only corresponding to scribble annotations; (2) local saliency coherence loss ($L_{lsc}$) to enforce similar pixels in the adjacent region to share consistent saliency scores; (3) smoothness loss  ($L_{sl}$) to guarantee the local smoothness and salient distinction along image edges; (4) structure consistency loss ($L_{ssc}$) to enforce the consistency on saliency maps from different input scales.

The scribble supervision loss $L_{scribble}$ is given by the combination of the above loss functions:
\begin{equation}
\begin{aligned}
L_{scribble}=L_{pce}+L_{lsc}+L_{sl}+L_{ssc}
\end{aligned}
\end{equation}
The loss uses the sparse annotations which indicate the foreground and background and adds the local and scale constraint. However, all the cues come from RGB images. When the quality of RGB images are low, the performance will be bad. Next, we will elaborate on the additional pseudo label supervision from RGB and thermal modalities.

\subsection{Pseudo Label Generation and Supervision}
Existing methods \cite{wei2021scribble,chen2022scribble2d5} generate expanded  labels by collecting the superpixels that scribbles pass through in RGB images. In the RGB-T SOD task, the expanded  labels can be obtained from RGB images and thermal images. From the observation of Fig \ref{fig:Dataset}, we find the inaccuracy and complementarity  of multi-modal expanded  labels.
Accordingly, we attempt to predict the pseudo labels from the two modalities via the supervision of expanded labels and then aggregate them to form the final pseudo labels, rather than directly use the expanded labels.

Our four-step solution includes expansion, prediction, aggregation, and supervision.

\textbf{Expansion.}
A foreground scribble expansion module is proposed to generate expanded labels of two modalities.
Specifically, the foreground scribbles are extracted from scribble annotations and the image is partitioned into multiple superpixels via SLIC.
If the intersection of a superpixel and foreground scribble is not null, the superpixel is marked as the foreground.
As a result, the scribble annotation $Y$ is expanded via the RGB superpixels and thermal superpixels, generating corresponding expanded RGB labels $\hat Y^x=\{\hat y^x_i\}_{i=1}^N$ and expanded thermal labels $\hat Y^t=\{\hat y^t_i\}_{i=1}^N$.%, respectively.
The algorithm of foreground scribble expansion module is shown in Alg  \ref{alg:expansion}.
\begin{algorithm}[!htp]

%\footnotesize
	\caption{Foreground Scribble Expansion.}
     \renewcommand{\algorithmicrequire}{\textbf{Input:}}
      \renewcommand{\algorithmicensure}{\textbf{Output:}}
	\label{alg:expansion}
	\begin{algorithmic}[1]

	\REQUIRE
        (1) An RGB or thermal image $M$; \\(2) Scribble annotation $Y=\{y_i\}_{i=1}^N$, where $y_i=1/2/0$ indicates foreground, background, and unknown pixels, respectively, and $N$ is the number of pixels;\\

    \ENSURE Expanded labels $\hat Y$;\\

 /*Step1: Use SLIC algorithm to segment the image $M$,  generating the superpixel image mask set $\mathbb{P}=\{P^k\}_{k=1}^T$, where $T$ is the number of superpixels; Each $P^k$ only highlights the current superpixel.*/ \\
            \STATE $\mathbb{P}\leftarrow SLIC(M)$; \\

         /*Step2: Extract the foreground scribble $\textit{FS}=\{\textit{fs}_i\}_{i=1}^N$ from scribble annotation $Y$.*/ \\
         \STATE Initialize $\textit{FS}\leftarrow zeros(N)$;

        \FOR{ $i$=1 to $N$}
            \IF{$y_i=1$ }
            \STATE $\textit{fs}_i\leftarrow$255

            \ELSE
            \STATE $\textit{fs}_i\leftarrow$0\\
            \ENDIF

        \ENDFOR\\
        /*Step3: Mark the superpixel as the foreground if the intersection of a superpixel and foreground scribble is not null.*/ \\
        \STATE Initialize $\hat{Y}\leftarrow zeros(N)$;
        \FOR{each $P^k$ in $\mathbb{P}$}
                \IF {$P^k \cap \textit{FS}$  $\neq\varnothing$ }
                    \STATE $\hat Y \leftarrow \hat Y$+$P^k$\\
                \ENDIF
        \ENDFOR
	\RETURN  $\hat Y$.
	\end{algorithmic}
\end{algorithm}

\begin{figure}[!htp]
  \centering
  \includegraphics[width=1\linewidth]{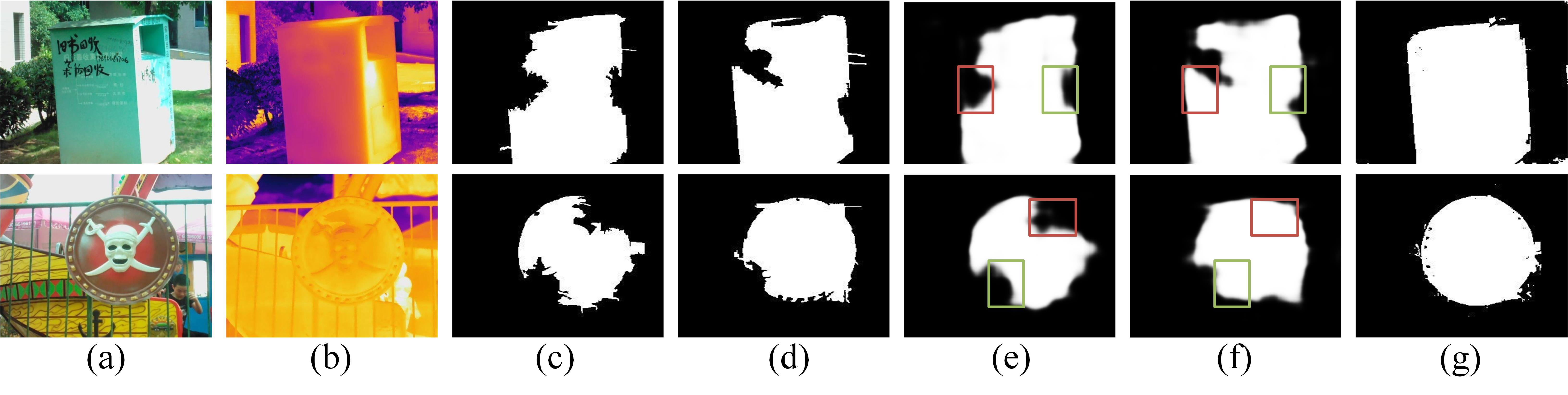}
  \caption{{Examples to show the advantage of predicted multi-modal labels compared with  expanded multi-modal labels and the complementarity of multi-modal labels. (a-b) RGB and thermal images; (c-d) expanded RGB and thermal labels; (e-f) predicted RGB and thermal labels; (g) aggregated pseudo labels.}}
  \label{fig:AblationExpandPredict}
\end{figure}

\textbf{Prediction.}
The expanded labels are obtained by SLIC which considers the local consistency.
However, there are some noises in the boundary of objects, which can be shown in Fig \ref{fig:AblationExpandPredict} (c-d).
Therefore, a prediction module is designed  to predict the pseudo labels from each modality under the supervision of the expanded multi-modal labels.
\begin{equation}
\begin{aligned}
P^x=\textit{Prediction}(f^x_4)\\
P^t=\textit{Prediction}(f^t_4)\\
\end{aligned}
\end{equation}
where $P^x=\{p^x_i\}_{i=1}^N$ and $P^t=\{p^t_i\}_{i=1}^N$ are predicted RGB and thermal labels.

The prediction module consists of successive convolution and upsampling operations conducted on the highest layer features of encoders, resulting more smoothed boundaries, as shown in Fig \ref{fig:AblationExpandPredict} (e-f)vs.(c-d). %In addition, the two encoders also learn the local constraint via expanded multi-modal labels.
%(MFNet)As is pointed out in [9, 4, 26, 8], the convolutional neural networks possess good robustness to noisy labels. Therefore, the inaccurate saliency cues in pseudo labels can be gradually corrected by the convolution layers in DF.
The predicted  labels are supervised by the expanded labels via cross entropy loss.
%\begin{equation}
%\begin{aligned}
%L_{ce}^{k}=-\frac{1}{ N}\sum_{i}^{}{p_i^{k}\log \hat y_i^{k}+(1-p_i^{k})\log (1-\hat y_i^{k})}\\
%\end{aligned}
%\end{equation}
\begin{equation}
\begin{aligned}
L_{ce}^{x}=-\frac{1}{ N}\sum_{i}^{}{p_i^{x}\log \hat y_i^{x}+(1-p_i^{x})\log (1-\hat y_i^{x})}\\
L_{ce}^{t}=-\frac{1}{ N}\sum_{i}^{}{p_i^{t}\log \hat y_i^{t}+(1-p_i^{t})\log (1-\hat y_i^{t})}\\
\end{aligned}
\end{equation}

\textbf{Aggregation.}
Two predicted labels are averaged and smoothed by original RGB images via PAR method \cite{ru2022learning} to generate aggregated pseudo labels $\bar Y=\{\bar y_i\}_{i=1}^N$.
PAR incorporates the RGB and spatial information to define an affinity kernel measuring its proximity to its neighbours
and iteratively applies this kernel to the average value via an adaptive convolution \cite{su2019pixel}.
%PAR takes the individual pixel in predicted pseudo map as seeds, and infers the scores of its neighbor pixels using the RGB appearance information.

\begin{equation}
\begin{aligned}
\bar Y=\textit{PAR}(\textit{Avg}(P^x,P^t),X)
\end{aligned}
\end{equation}
Here, we only use RGB images instead of the thermal images to refine the average values because RGB images provide the more abundant information benefiting the refinement compared with thermal images.

From Fig \ref{fig:AblationExpandPredict} (g)vs.(e-f), we can observe that the aggregated pseudo labels are better via the mutual supplementation of two modalities.

\textbf{Supervision.}
The supervision of aggregated pseudo labels are  appended to the baseline by pixel position aware loss $L_{ppa}$.
%\vspace{-1.5em}
\begin{equation}
\begin{aligned}
L_{ppa}=L_{wbce}(R,\bar Y)+L_{wiou}(R,\bar Y)
\end{aligned}
\end{equation}
where $L_{wbce}$ and $L_{wiou}$ are the weighted binary cross entropy loss and  weighted IoU loss, respectively \cite{wei2020f3net}.

The final loss function $L$ for training the proposed model is given by:

\begin{equation}
\begin{aligned}
L=L_{scribble}+L_{ce}^x+L_{ce}^t+L_{ppa}
\end{aligned}
\end{equation}

During testing, we only retain the baseline network and discard the prediction modules and foreground scribble expansion modules for acceleration.
\section{Experiments}
\subsection{Dataset and Evaluation Metrics}
%\textbf{RGB-T dataset.}
We relabel RGB-T SOD training set via scribbles, which includes 2,500 image pairs in VT5000 \cite{tu2020rgbt}, namely RGBT-S dataset. The remaining samples of VT5000, the whole VT821 \cite{wang2018rgb} and VT1000 \cite{tu2019rgb} are used to be tested.
VT821\cite{wang2018rgb} contains 821 manually registered image pairs, so the quality of some thermal infrared images are bad (vacant regions or grey-scale map).
VT1000\cite{tu2019rgb} contains 1,000 RGB-T image pairs captured with highly aligned RGB and thermal cameras.
VT5000\cite{tu2020rgbt} contains 5,000 pairs of high-resolution, high-diversity, and low-deviation RGB-T images.
The evaluation metrics follow the BBSNet \cite{fan2020bbs} and SwinNet \cite{liu2021swinnet}.

\subsection{Implementation Details}
Our model is implemented based on PyTorch on a PC with
a NVIDIA RTX 3090 GPU.
The input image size is $256\times256$. The number of superpixels is $40$.
Flipping, cropping, and rotating are used to augment the training set.
We employ the Adam optimizer %\cite{kingma2014adam}
to train our model.
The max training epoch is set to 300 and  batch size is set to 16.
The initial learning rate is 5e-5 and the learning rate will be divided by 10 every 200 epochs. The train process takes about 20 hours.
\subsection{Comparison Experiments}
\textbf{Comparison Methods and Instructions.}
The fully-supervised multi-modal methods (BBSNet \cite{fan2020bbs},  MIDD \cite{tu2021multiinteractive},
SwinNet \cite{liu2021swinnet}, TNet \cite{cong2022does}),
scribble-supervised RGB methods (
WSSA \cite{zhang2020weakly},
SCWSSOD \cite{yu2021structure},
%MFNet\cite{piao2021mfnet},
SCOD \cite{he2022weakly}), scribble-supervised RGB-D   method (DENet \cite{xu2022weakly}), and the video  ones (WSVOD \cite{zhao2021weakly}) are compared.
For fairness, we use the source codes provided by the authors to re-train the RGBT-S datasets. For RGB based methods, multi-modal image pairs are first concatenated and then fed into the models. %Our method provides ResNet-50 \cite{he2016deep} and PVTv2-B2 \cite{wang2021pyramid} version.
Our method provides ResNet-50\cite{he2016deep} and PVTv2-B2\cite{wang2021pyramid} version. %ResNet is ResNet-50  and PVT is PVTv2-B2 .
All the evaluation metrics are calculated using the same evaluation code on result saliency maps.
The win/tie/losses of fully-supervised methods show our PVT version vs. the other fully-supervised methods, while those of scribble-supervised methods mean our ResNet version vs. the other scribble-supervised ones.

\textbf{Quantitative Analysis.}
From Table \ref{tab:Compare}, we can find the below facts.
Our ResNet model achieves the best results among the scribble-supervised methods. It fairly verifies the advantage of our model.
Furthermore, our PVT model significantly outperforms the other scribble-supervised methods. Moreover, it is also comparable with fully-supervised methods which can be seen from their win/tie/loss values. It scientifically  exhibits the superiority of the proposed model. The experimental data discloses the effectiveness of superpixel expansion of scribble annotations because the superpixel can provide a coarse boundary based on the given scribble. Especially, two-modal superpixel expansion can complement each other and play an approximate role as ground truth. Certainly, the proposed prediction module also polishes the coarse boundary of SLIC segmentation. It also improves the performance.
\begin{table*}[!htp]
%\vspace{-1em}
\caption{Comparison with fully-supervised methods and scribble-supervised methods. The best results in each group are in bold.}
\centering
\resizebox{0.95\linewidth}{!}
{
   \begin{tabular}{c|c|cccc|cccccccc}
  \hline\toprule
    \multirow{3}{*}{Datasets}& \multirow{3}{*}{Metric} &\multicolumn{4}{c|}{\centering Fully Sup. Methods}
    &\multicolumn{7}{c}{\centering Scribble Sup. Methods}\\

     \cline{3-13}

    &&BBSNet \cite{fan2020bbs}&MIDD \cite{tu2021multiinteractive} &SwinNet \cite{liu2021swinnet}&TNet \cite{cong2022does}&WSSA \cite{zhang2020weakly}&SCWSSOD \cite{yu2021structure}&SCOD \cite{he2022weakly} & DENet \cite{xu2022weakly}& WSVOD \cite{zhao2021weakly} &\multicolumn{2}{c}{\centering Ours}\\
  &&ECCV20&TIP21 &TCSVT22&TMM22&CVPR20&AAAI21&AAAI23 & TIP22&CVPR21& ResNet&PVT\\

  \midrule

  \multirow{4}{*}{\textit{VT821} \cite{wang2018rgb}}
    & $S\uparrow$    &.862  &.871   & \textbf{.904}&.899& .805  &.845    &.822  &.813&.822&.857&$\textbf{.895}$\\

    & $F_\beta \uparrow$      &.762  &.804  &\textbf{.847} &.842&.722 &.802    &.772 &.705& .755&.835 &$\textbf{.875}$\\

    & $E_{\xi}\uparrow$      &.879  &.895  &\textbf{.926}&.919 &.848 &.884   &.879&.827& .868&.904 &$\textbf{.934}$ \\

    & $M\downarrow$        &.046 &.045  & \textbf{.030}&\textbf{.030}&.058 &.050   &.048&.054&.052 &.036 &$\textbf{.027}$ \\
        \cline{1-13}

  \multirow{4}{*}{\textit{VT1000} \cite{tu2019rgb}}
    & $S\uparrow$   &.926 &.915  & \textbf{.938}&.929&.882 &.907   &.880&.906& .886&.906 &$\textbf{.925}$\\

    & $F_\beta \uparrow$    &.847 &.882  & \textbf{.896}&.889& .842& .895  &.849&.870&.863&.899&$\textbf{.914}$  \\

    & $E_{\xi}\uparrow$    &.920 &.933  & \textbf{.947}&.937&.907 &.937   &.920&.925&.916&.939 &$\textbf{.949}$ \\

    & $M\downarrow$        &.027 &.027  & \textbf{.018}&.021&.036 &.027   &.035&.028& .035&.027 &$\textbf{.020}$\\
      \cline{1-13}

  \multirow{4}{*}{\textit{VT5000} \cite{tu2020rgbt}}
    & $S\uparrow$   &.873 &.867  &\textbf{.912}&.895 &.820 &.839   &.827&.839&.812 &.843 &$\textbf{.877}$\\

    & $F_\beta \uparrow$  &.776 &.801  &\textbf{.865} &.846&.747 &.801   &.771&.769&.752 &.817 &$\textbf{.858}$ \\

    & $E_{\xi}\uparrow$   &.889 &.897  &\textbf{.942} &.927&.875 &.891   &.890&.883&.873 &.903 &$\textbf{.913}$ \\

    & $M\downarrow$     &.044 &.043  &\textbf{.026 }&.033&.054 &.048   &.049&.048&.055 &.042 &$\textbf{.033}$ \\

\midrule
  \multicolumn{2}{c|}{\centering Ours vs. methods win/tie/loss}
    &11/0/1 &12/0/0 &5/0/7 &7/1/4 &12/0/0 &10/1/1 &12/0/0 &11/1/0 &12/0/0 &/ &/
  \\

  \bottomrule
  \hline
  \end{tabular}
}
\label{tab:Compare}
\end{table*}

\begin{figure}[!htp]

  \centering
  \includegraphics[width=1\linewidth]{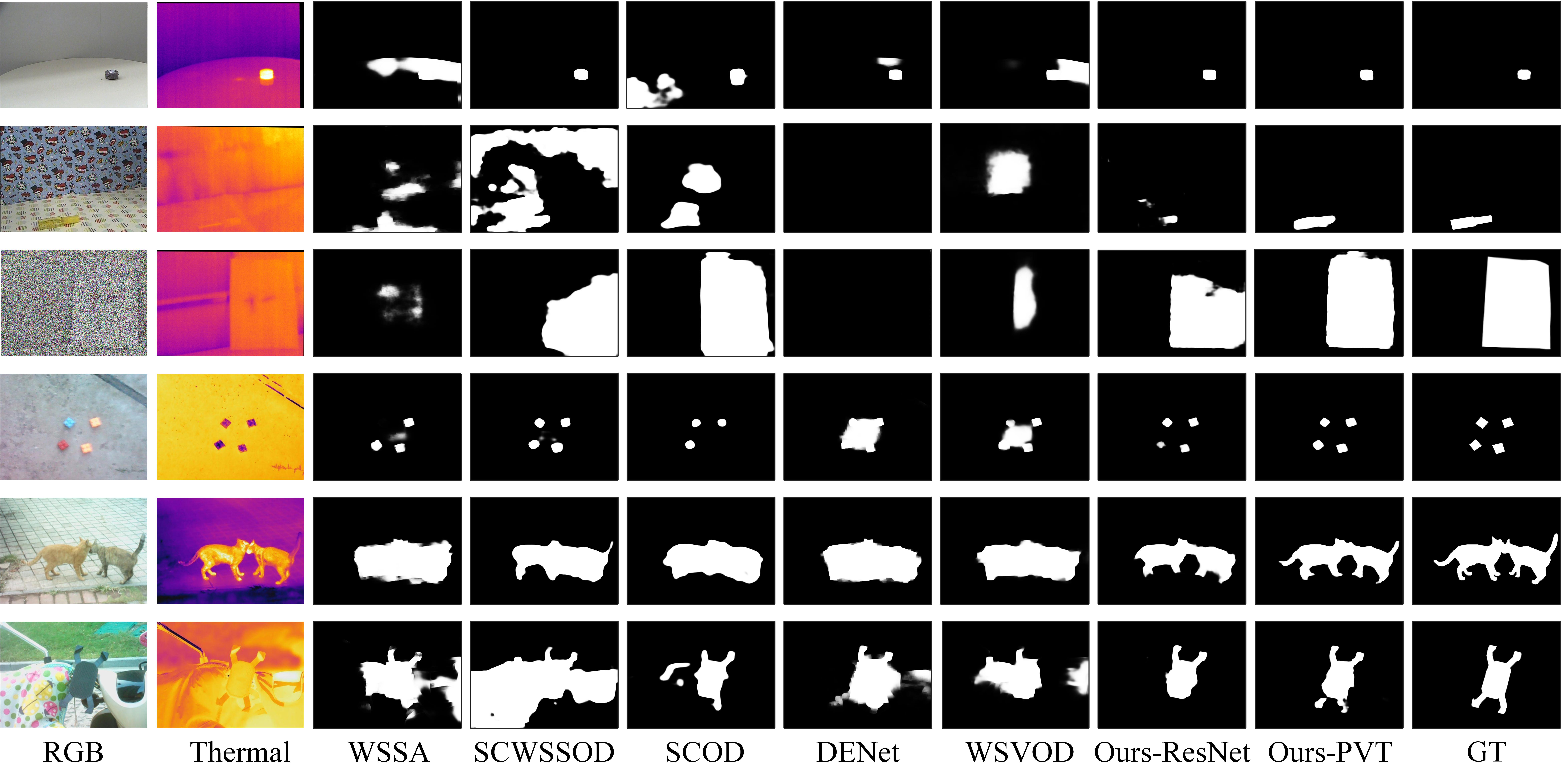}\\
  \caption{{Visual performance of scribble-supervised compared methods.}
  \label{fig:Visual}}
\end{figure}

\begin{figure}[!htp]
%\vspace{-1.0em}
  \centering
  \includegraphics[width=1\linewidth]{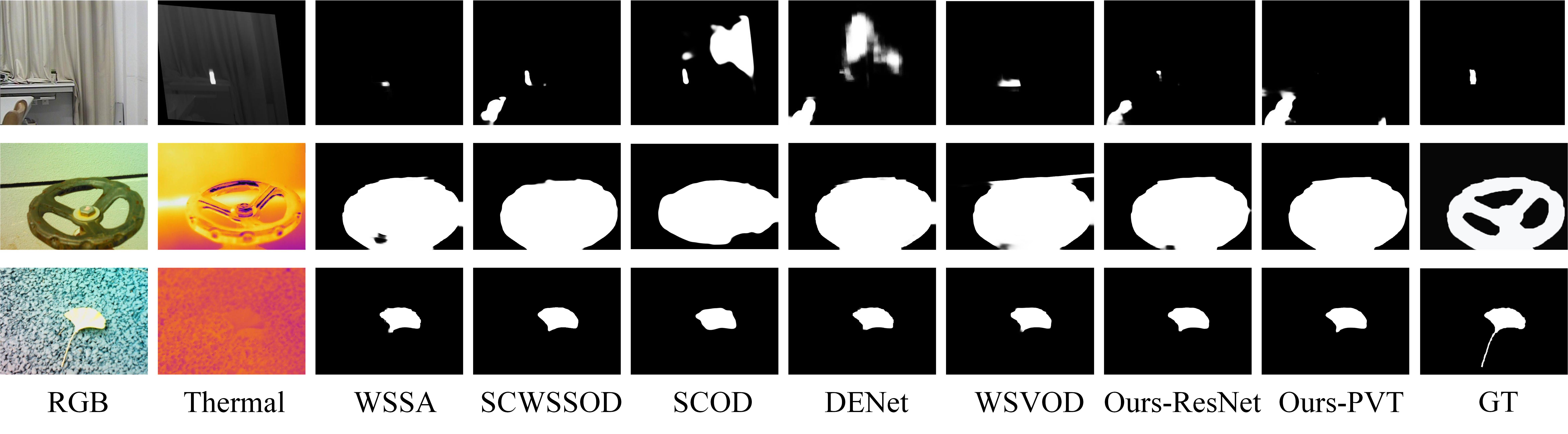}
  \caption{{Failure cases.}
  \label{fig:Failure}}
\end{figure}

\textbf{Visual Analysis.}
%对比图表示我们的方法在各种具有挑战性的场景中表现良好，包括了微小物体（第一行），复杂背景（第二行），RGB图片加噪模糊（第三行），小且多物体（第四行），显著目标连通（第五行），图片内容复杂（第六行）
Fig \ref{fig:Visual} shows some visual examples in small objects, complex background, blur RGB images, multiple objects, connected objects, and cluttered scenes. Our results are obviously better, which indicate the advantage of our model. Certainly, there are some failure cases, for example, transparent, hollow, and thin objects shown in Fig \ref{fig:Failure}. It can be inferred from the performances of the other scribble-supervised methods that these scenes will be the great  challenge of future research.
%第一行显著物体是个透明的杯子，由于物体是透明的且与背景颜色太过相似，RGB依据颜色进行超像素分割的时候就会产生较大误差导致预测不准。第二行是个镂空的方向盘，网络在预测的时候就很容易将整个物体进行预测，不容易分出镂空部分。第三行是一个带有细枝干的银杏叶，我们的网络虽然能够检测出主体部分，但对细小的其余部位未能准确识别

\subsection{Ablation Study}
\textbf{Expanded vs. Predicted}
Ablation study about expanded and predicted labels are shown in Table \ref{tab:ExpandedPseudoLabelAblation}. The performance under the appended supervision of aggregated pseudo labels which are averaged and smoothed by expanded multi-modal labels is better than the baseline. It shows the effectiveness of scribble expansion via superpixels. Furthermore, when expanded labels are replaced with the predicted ones, the performance is improved. It discloses the fact that prediction module plays an important role in polishing the boundary noises of expanded labels and preventing  model from consistently learning false signals, generating the more accurate result.

\begin{table}[!htp]
%\vspace{-1.0em}
\caption{Ablation study about expanded labels and predicted labels. %No.1: baseline; No.2: baseline+direct expanded  labels; No.3: baseline+predicted expanded labels
}
\centering
\resizebox{1\linewidth}{!}
{
      \begin{tabular}{c|cccc|cccc|cccc}
    \toprule

   \multirow{2}{*}{\centering Models} & \multicolumn{4}{c|}{\centering VT821 \cite{wang2018rgb}} & \multicolumn{4}{c|}{\centering VT1000 \cite{tu2019rgb}} & \multicolumn{4}{c}{\centering VT5000 \cite{tu2020rgbt}}  \\
      &$S$$\uparrow$& $F_{\beta}$$\uparrow$& $E_{\xi}$ $\uparrow$& $M\downarrow$
         &$S$$\uparrow$& $F_{\beta}$$\uparrow$& $E_{\xi}$ $\uparrow$& $M\downarrow$
         &$S$$\uparrow$& $F_{\beta}$$\uparrow$& $E_{\xi}$ $\uparrow$& $M\downarrow$
      \\
    \hline
Baseline
     &.873&.860&.921&.032
      &.914&.903&.937&.025
      &.858&.844&.914&.041
      \\
+Expanded  &.889&.863&.928&.030
      &.919&.903&.941&.024
      &.873&.848&.926&.036
      \\
+Predicted   &\textbf{.895}&\textbf{.875}&\textbf{.934}&\textbf{.027}
      &\textbf{.925}&\textbf{.914}&\textbf{.949}&\textbf{.020}
      &\textbf{.877}&\textbf{.858}&\textbf{.931}&\textbf{.033}
      \\

    \bottomrule
\end{tabular}
}
\label{tab:ExpandedPseudoLabelAblation}
\end{table}

\textbf{Multi-modal vs. Single-modal}
Ablation study about multi-modal supervision on two prediction branches is shown in Table \ref{tab:MultimodalAblation}. We can observe that multi-modal labels outperforms single-modal labels because the last two rows are better than the first two rows. Meanwhile, it is better to supervise each  branch with the corresponding modality from the comparison between the last two rows.
It is inferred from the data that the encoder can learn the local cue from superpixel segmentation via the corresponding multi-modal supervision.%which provides more boundary cues

\begin{table}[!htp]
%\vspace{-1.0em}
\caption{Ablation study about the supervision of two prediction modules. ``X+Y": X supervises the RGB branch and Y supervises the thermal branch, respectively.} %No.1: multi-modal supervision (RGB and thermal branch use expanded  labels of each modality); No.2: single-modal supervision (RGB and thermal branch use expanded  labels of RGB modality); No.3: single-modal supervision (RGB and thermal branch use expanded  labels of thermal modality)

\centering
\resizebox{1\linewidth}{!}
{
      \begin{tabular}{c|cccc|cccc|cccc}
    \toprule

   \multirow{2}{*}{\centering Models} & \multicolumn{4}{c|}{\centering VT821 \cite{wang2018rgb}} & \multicolumn{4}{c|}{\centering VT1000 \cite{tu2019rgb}} & \multicolumn{4}{c}{\centering VT5000 \cite{tu2020rgbt}}  \\
      &$S$$\uparrow$& $F_{\beta}$$\uparrow$& $E_{\xi}$ $\uparrow$& $M\downarrow$
         &$S$$\uparrow$& $F_{\beta}$$\uparrow$& $E_{\xi}$ $\uparrow$& $M\downarrow$
         &$S$$\uparrow$& $F_{\beta}$$\uparrow$& $E_{\xi}$ $\uparrow$& $M\downarrow$
      \\
    \hline
RGB+RGB
     &.888&.866&.925&.028
      &.923&\textbf{.915}&.948&.021
      &.872&.848&.922&.035
      \\
Thermal+Thermal  &.884&.869&.930&.029
      &.919&.910&.946&.023
      &.872&.849&.927&.035
      \\
Thermal+RGB
      &.890&.871&\textbf{.935}&.028
      &.922&.906&.944&.021
      &\textbf{.877}&.852&.928&.034
      \\
RGB+Thermal   &\textbf{.895}&\textbf{.875}&.934&\textbf{.027}
      &\textbf{.925}&.914&\textbf{.949}&\textbf{.020}
      &\textbf{.877}&\textbf{.858}&\textbf{.931}&\textbf{.033}
      \\
    \bottomrule
\end{tabular}
}
\label{tab:MultimodalAblation}
\end{table}

\textbf{Scribble  vs. Aggregated Pseudo Label}
Ablation study about scribble supervision and aggregated pseudo label ones is shown in Table \ref{tab:ScribblePseudoLabelAblation}. When our model removes the original scribble supervision, the performance is reduced from the comparison between the 1$^{st}$ and 3$^{rd}$ rows. It demonstrates that aggregated pseudo labels are  inaccurate. So we need to continue to restrain the model via scribble supervision so that the model can better use the information provided by scribble annotation without being affected by inaccurate pseudo labels. Of course, it is not enough to supervise the model only using  sparse scribble annotations, which can be seen from the comparison between 2$^{nd}$ and 3$^{rd}$ rows. %Because the scribble annotations only come from RGB images, the thermal images can play a  complementary role in the scenes where the qualities of RGB images are low. %In our model, the generation of pseudo labels combines the information of RGB and thermal modalities.
Scribble annotations which only utilize RGB information can't provide enough constraint when RGB images are not good. The additional pseudo label supervision can improve the performance.

\begin{table}[!htp]
%\vspace{-1.0em}
\caption{Ablation study about the supervision of scribble and aggregated pseudo labels.} %No.1: multi-modal supervision (RGB and thermal branch use expanded  labels of each modality); No.2: single-modal supervision (RGB and thermal branch use expanded  labels of RGB modality); No.3: single-modal supervision (RGB and thermal branch use expanded  labels of thermal modality)

\centering
\resizebox{1\linewidth}{!}
{
       \begin{tabular}{c|cccc|cccc|cccc}
    \toprule

   \multirow{2}{*}{\centering Models} & \multicolumn{4}{c|}{\centering VT821 \cite{wang2018rgb}} & \multicolumn{4}{c|}{\centering VT1000 \cite{tu2019rgb}} & \multicolumn{4}{c}{\centering VT5000 \cite{tu2020rgbt}}  \\
      &$S$$\uparrow$& $F_{\beta}$$\uparrow$& $E_{\xi}$ $\uparrow$& $M\downarrow$
         &$S$$\uparrow$& $F_{\beta}$$\uparrow$& $E_{\xi}$ $\uparrow$& $M\downarrow$
         &$S$$\uparrow$& $F_{\beta}$$\uparrow$& $E_{\xi}$ $\uparrow$& $M\downarrow$
      \\
  \hline

w/o scribble label
     &.872&.820&.911&.034
      &.903&.879&.936&.028
      &.855&.789&.904&.040
      \\
w/o pseudo label%baseline
   &.873&.860&.921&.032
      &.914&.903&.937&.025
      &.858&.844&.914&.041
      \\
Ours   &\textbf{.895}&\textbf{.875}&\textbf{.934}&\textbf{.027}
      &\textbf{.925}&\textbf{.914}&\textbf{.949}&\textbf{.020}
      &\textbf{.877}&\textbf{.858}&\textbf{.931}&\textbf{.033}
      \\
    \bottomrule
\end{tabular}
}
\label{tab:ScribblePseudoLabelAblation}
\end{table}
\subsection{Model Generality Analysis}
The model can be applied to RGB-D and video scribble-supervised SOD tasks.

In the experiments, RGB-D DENet \cite{xu2022weakly} and video WSVOD \cite{zhao2021weakly} are retrained in both RGB-D and video scribble-supervised SOD datasets.
In the top and bottom of Table \ref{tab:ExpandCompare}, six RGB-D and six video SOD datasets are tested, respectively.
From the result, we find our ResNet version wins 20 out of 24 compared with DENet in the RGB-D task and thoroughly defeat WSVOD. Meanwhile, in the video task, compared with WSVOD and DENet, our model wins out in most evaluation metrics.
%ours wins 13 out of 24  and  16 out of 24.
It is very impressive  that our PVT version achieves the state-of-the-art performance. It verifies the generality of our model. It is first time to use a unified scribble-supervised model to run RGB-T, RGB-D, and video scribble-supervised SOD dataset.
By the analysis, we find DENet and WSVOD only use multi-modal fusion in model structures and don't excavate the multi-modal cues in the supervision. Our model is the consistently excellent because we innovatively play the role of the other modality in the supervision.

\begin{table*}[htp!]
%\vspace{-1.0em}
\caption{Comparison with  scribble-supervised multi-modal SOD methods in RGB-D and video datasets. {\textcolor{red}{Red}}: the best. {\textcolor{blue}{Blue}}: the second best.}%RGB-D DENet and video WSVOD in RGB-D and video}
\centering
\resizebox{\linewidth}{!}
{
   \begin{tabular}{c|cccc|cccc|cccc|cccc|cccc|cccc}
  \hline\toprule

  \multirow{2}{*}{\centering RGB-D SOD}&\multicolumn{4}{c|}{\centering NJU2K\cite{ju2014depth}} & \multicolumn{4}{c|}{\centering STERE\cite{niu2012leveraging}} & \multicolumn{4}{c|}{\centering DES\cite{cheng2014depth}}&\multicolumn{4}{c|}{\centering NLPR\cite{peng2014rgbd}} & \multicolumn{4}{c|}{\centering LFSD\cite{li2014saliency}} & \multicolumn{4}{c}{\centering SIP\cite{fan2020rethinking}} \\

  %\multirow{2}{*}{\centering RGB-D SOD}&\multicolumn{4}{c|}{\centering NJU2K} & \multicolumn{4}{c|}{\centering STERE} & \multicolumn{4}{c|}{\centering DES}&\multicolumn{4}{c|}{\centering NLPR} & \multicolumn{4}{c|}{\centering LFSD} & \multicolumn{4}{c}{\centering SIP} \\

      &$S\uparrow$ &$F_\beta\uparrow$ &$E_{\xi}\uparrow$&M$\downarrow$
   &$S\uparrow$ &$F_\beta\uparrow$ &$E_{\xi}\uparrow$&M$\downarrow$
    &$S\uparrow$ &$F_\beta\uparrow$ &$E_{\xi}\uparrow$&M$\downarrow$
    &$S\uparrow$ &$F_\beta\uparrow$ &$E_{\xi}\uparrow$&M$\downarrow$
   &$S\uparrow$ &$F_\beta\uparrow$ &$E_{\xi}\uparrow$&M$\downarrow$
    &$S\uparrow$ &$F_\beta\uparrow$ &$E_{\xi}\uparrow$&M$\downarrow$\\
 \midrule
  DENet(TIP22) \cite{xu2022weakly}

    & $\textbf{\textcolor{blue}{.875}}$ & $\textbf{\textcolor{blue}{.861}}$ & .902 & $\textbf{\textcolor{blue}{.053}}$
    & .874 & .847 & .910 & .051
    & .898 & .882 & .954 & $\textbf{\textcolor{blue}{.029}}$
    & $\textbf{\textcolor{blue}{.898}}$ & .859 & .944 & .032
    & .827 & .820 & .866 & .085
    & .846 & .827 & .905 & .063 \\
  WSVOD(CVPR21) \cite{zhao2021weakly}

    & .814 & .804 & .861 & .078
    & .835 & .823 & .887 & .066
    & .844 & .826 & .908 & .043
    & .863 & .833 & .927 & .041
    & .751 & .744 & .782 & .136
    & .759 & .728 & .849 & .105 \\
  %DENet2

  %  &.877&.860&.904&.051&.876&.852&.913&.048 &.885 &.874 &.941 &.030 &.900 &.862 &.946 &.031 &.826 &.819 &.867 &.086 &.846 &.826 &.905 &.063\\

  %\cline{1-25}
  Ours-ResNet
  & .859 & .860 & $\textbf{\textcolor{blue}{.903}}$ & .058
  & $\textbf{\textcolor{blue}{.875}}$ & $\textbf{\textcolor{blue}{.873}}$ & $\textbf{\textcolor{blue}{.923}}$ & $\textbf{\textcolor{blue}{.047}}$
  & $\textbf{\textcolor{blue}{.905}}$ & $\textbf{\textcolor{red}{.902}}$ & $\textbf{\textcolor{blue}{.959}}$ & $\textbf{\textcolor{red}{.023}}$
  & .897 & $\textbf{\textcolor{blue}{.879}}$ & $\textbf{\textcolor{blue}{.947}}$ & $\textbf{\textcolor{blue}{.029}}$
  & $\textbf{\textcolor{blue}{.838}}$ & $\textbf{\textcolor{blue}{.849}}$ & $\textbf{\textcolor{blue}{.882}}$ & $\textbf{\textcolor{blue}{.076 }}$
  & $\textbf{\textcolor{blue}{.851}}$ & $\textbf{\textcolor{blue}{.846}}$ & $\textbf{\textcolor{blue}{.912}}$ &$\textbf{\textcolor{blue}{.059}}$
  \\

  Ours-PVT
  & $\textbf{\textcolor{red}{.885}}$ & $\textbf{\textcolor{red}{.886}}$ & $\textbf{\textcolor{red}{.925}}$ & $\textbf{\textcolor{red}{.044}}$
  & $\textbf{\textcolor{red}{.887}}$ & $\textbf{\textcolor{red}{.876}}$ & $\textbf{\textcolor{red}{.929}}$ & $\textbf{\textcolor{red}{.042}}$
  & $\textbf{\textcolor{red}{.913}}$ & $\textbf{\textcolor{blue}{.897}}$ & $\textbf{\textcolor{red}{.961}}$ & $\textbf{\textcolor{red}{.023}}$
  & $\textbf{\textcolor{red}{.906}}$ & $\textbf{\textcolor{red}{.884}}$ & $\textbf{\textcolor{red}{.958}}$ & $\textbf{\textcolor{red}{.024}}$
  & $\textbf{\textcolor{red}{.853}}$ & $\textbf{\textcolor{red}{.862}}$ & $\textbf{\textcolor{red}{.901}}$ & $\textbf{\textcolor{red}{.068}}$
  & $\textbf{\textcolor{red}{.855}}$ & $\textbf{\textcolor{red}{.851}}$ & $\textbf{\textcolor{red}{.916}}$ & $\textbf{\textcolor{red}{.056}}$
  \\
\midrule
\midrule
\multirow{2}{*}{\centering Video SOD}&\multicolumn{4}{c|}{\centering SegV2\cite{li2013video}
} & \multicolumn{4}{c|}{\centering DAVSOD \cite{fan2019shifting}
} & \multicolumn{4}{c|}{\centering VOS \cite{li2017benchmark}
}&\multicolumn{4}{c|}{\centering FBMS \cite{ochs2013segmentation}} & \multicolumn{4}{c|}{\centering DAVIS \cite{perazzi2016benchmark}} & \multicolumn{4}{c}{\centering ViSal \cite{wang2015consistent} }\\

      &$S\uparrow$ &$F_\beta\uparrow$ &$E_{\xi}\uparrow$&M$\downarrow$
   &$S\uparrow$ &$F_\beta\uparrow$ &$E_{\xi}\uparrow$&M$\downarrow$
    &$S\uparrow$ &$F_\beta\uparrow$ &$E_{\xi}\uparrow$&M$\downarrow$
    &$S\uparrow$ &$F_\beta\uparrow$ &$E_{\xi}\uparrow$&M$\downarrow$
   &$S\uparrow$ &$F_\beta\uparrow$ &$E_{\xi}\uparrow$&M$\downarrow$
    &$S\uparrow$ &$F_\beta\uparrow$ &$E_{\xi}\uparrow$&M$\downarrow$\\
\midrule
  WSVOD(CVPR21) \cite{zhao2021weakly}

    & $\textbf{\textcolor{blue}{.785}}$ & .685 & $\textbf{\textcolor{blue}{.866}}$ & .032
    & $\textbf{\textcolor{blue}{.697}}$ & .564 & .743 & .101
    & $\textbf{\textcolor{blue}{.713}}$ & $\textbf{\textcolor{blue}{.607}}$ & $\textbf{\textcolor{blue}{.743}}$ & $\textbf{\textcolor{blue}{.085}}$
    & $\textbf{\textcolor{blue}{.663}}$ & .650 & $\textbf{\textcolor{blue}{.736}}$ & $\textbf{\textcolor{blue}{.064}}$
    & $\textbf{\textcolor{blue}{.814}}$ & .712 & .879 & .037
    & .766 & .706 & .834 & .036 \\
  DENet(TIP22) \cite{xu2022weakly}

    & .746 & .585 & .807 & .045
    & .675 & .513 & .712 & .113
    & .692 & .574 & .715 & .095
    & .652 & .589 & .706 & .078
    & .788 & .662 & .844 & .044
    & .746 & .655 & .807 & .045 \\
  Ours-ResNet

    & .741 & $\textbf{\textcolor{blue}{.725}}$ & .821 & $\textbf{\textcolor{blue}{.028}}$
    & .673 & $\textbf{\textcolor{blue}{.587}}$ & $\textbf{\textcolor{blue}{.752}}$ &$\textbf{\textcolor{blue}{.088}}$
    & .615 & .570 & .700 & .096
    & .650 & $\textbf{\textcolor{blue}{.670}}$ & .717 & .070
    & .808 & $\textbf{\textcolor{blue}{.789}}$ & $\textbf{\textcolor{blue}{.907}}$ & $\textbf{\textcolor{blue}{.033}}$
    & $\textbf{\textcolor{blue}{.781}}$ & $\textbf{\textcolor{blue}{.793}}$ & $\textbf{\textcolor{red}{.846}}$ & $\textbf{\textcolor{blue}{.024}}$ \\
  Ours-PVT

    & $\textbf{\textcolor{red}{.791}}$ & $\textbf{\textcolor{red}{.792}}$ & $\textbf{\textcolor{red}{.881}}$ & $\textbf{\textcolor{red}{.023}}$
    & $\textbf{\textcolor{red}{.723}}$ & $\textbf{\textcolor{red}{.657}}$ & $\textbf{\textcolor{red}{.789}}$ & $\textbf{\textcolor{red}{.077}}$
    & $\textbf{\textcolor{red}{.757}}$ & $\textbf{\textcolor{red}{.722}}$ & $\textbf{\textcolor{red}{.811}}$ & $\textbf{\textcolor{red}{.056}}$
    & $\textbf{\textcolor{red}{.691}}$ & $\textbf{\textcolor{red}{.715}}$ & $\textbf{\textcolor{red}{.745}}$ & $\textbf{\textcolor{red}{.048}}$
    & $\textbf{\textcolor{red}{.836}}$ & $\textbf{\textcolor{red}{.827}}$ & $\textbf{\textcolor{red}{.934}}$ & $\textbf{\textcolor{red}{.027}}$
    & $\textbf{\textcolor{red}{.785}}$ & $\textbf{\textcolor{red}{.798}}$ & $\textbf{\textcolor{blue}{.844}}$ & $\textbf{\textcolor{red}{.022}}$ \\
    \bottomrule
    \hline
  \end{tabular}
}
\label{tab:ExpandCompare}
\end{table*}
\section{Conclusion}
In the paper, a scribble-supervised RGB-T SOD model is first proposed based on relabeled RGBT-S dataset. The sparse scribble annotation is expanded via superpixel segmentation in RGB and thermal images. Due to coarse boundary of the expanded label, a prediction module is used to predict the pseudo labels. The predicted multi-modal pseudo labels are aggregated to supervise the model training. The model is also applied to RGB-D and video SOD tasks and achieves the state-of-the-art performance.
In the future, we will explore the scribble expansion solution via multi-modal supervoxels, in which the latent 3D geometry information will be further considered to improve the performance.

%\bibliographystyle{IEEEtran}
%\bibliography{RGBTScribblebib}

\end{document}